\documentclass[conference]{IEEEtran}
\IEEEoverridecommandlockouts
\usepackage{cite}
\usepackage{amsmath,amssymb,amsfonts}
\usepackage{algorithmic}
\usepackage{graphicx}
\usepackage{textcomp}
\usepackage{xcolor}

\usepackage{algorithm}
\usepackage{subcaption}
\usepackage[pagebackref=true,breaklinks=true,letterpaper=true,colorlinks,bookmarks=false]{hyperref}
\newtheorem{theorem}{Theorem}

\def\BibTeX{{\rm B\kern-.05em{\sc i\kern-.025em b}\kern-.08em
    T\kern-.1667em\lower.7ex\hbox{E}\kern-.125emX}}

\begin{document}

\title{Convolutional Feature Noise Reduction for 2D Cardiac MR Image Segmentation}

\onecolumn

\author {
    Hong Zheng\textsuperscript{\rm {1,2}}, Nan Mu\textsuperscript{\rm {1,3}}, Han Su\textsuperscript{\rm {1}}, Lin Feng\textsuperscript{\rm {1}}, Xiaoning Li\textsuperscript{\rm {*,1,3}}\\
    \textsuperscript{\rm 1}College of Computer Science, Sichuan Normal University, Chengdu, 610101, China\\
    \textsuperscript{\rm 2}School of Computing and Artificial Intelligence, Southwest Jiaotong University, Chengdu, 611756, China\\
    \textsuperscript{\rm 3}Visual Computing and Virtual Reality Key Laboratory of Sichuan Province, Chengdu, 610066, China\\
    \textsuperscript{\rm *}lxn@sicnu.edu.cn
    \thanks{\copyright~2025 IEEE. Author preprint. Accepted by IEEE International Conference on Multimedia \& Expo 2025, but not published due to no-show. }
}

\maketitle

\begin{abstract}
Noise reduction constitutes a crucial operation within Digital Signal Processing. Regrettably, it frequently remains neglected when dealing with the processing of convolutional features in segmentation networks. This oversight could trigger the butterfly effect, impairing the subsequent outcomes within the entire feature system. To complete this void, we consider convolutional features following Gaussian distributions as feature signal matrices and then present a simple and effective feature filter in this study. The proposed filter is fundamentally a low-amplitude pass filter primarily aimed at minimizing noise in feature signal inputs and is named Convolutional Feature Filter (CFF). We conducted experiments on two established 2D segmentation networks and two public cardiac MR image datasets to validate the effectiveness of the CFF, and the experimental findings demonstrated a decrease in noise within the feature signal matrices. To enable a numerical observation and analysis of this reduction, we developed a binarization equation to calculate the information entropy of feature signals. 
\end{abstract}

\section{Introduction}
\label{sec1}

Noise reduction, a fundamental operation in Digital Signal Processing (DSP)~\cite{b14}, aims to minimize the impact of extraneous information in the inputs on the subsequent processing results, leading to more dependable outcomes. For instance, a common approach involves using Gaussian Blur to smooth an image, resulting in clearer borders through convolution with the Laplacian kernel. In deep learning-based cardiac MR image (CMRI) segmentation studies, a typical emphasis lies in addressing noise within input images rather than considering the reduction of noise within the convolutional features. This is because most studies assume (1) noise reduction in inputs is sufficient for further processing, and (2) certain information in convolutional features, generated through non-linear convolution and activation operations, is missing. Consequently, to the best of our knowledge, most studies propose methods for feature enhancement or complementation, such as attention mechanisms~\cite{b10, b11} or anatomical shape priors~\cite{b6, b7}. However, according to information theory~\cite{b9}, any data processing method can act as a double-edged sword, potentially introducing redundancy while discarding relevant information. Therefore, it is plausible to hypothesize that noise exists within convolutional features, and the noise may adversely affect the subsequent results across the entire feature system, leading to suboptimal segmentation predictions for the networks. 

\begin{figure}[t]
\centering
\includegraphics[width=0.50\linewidth]{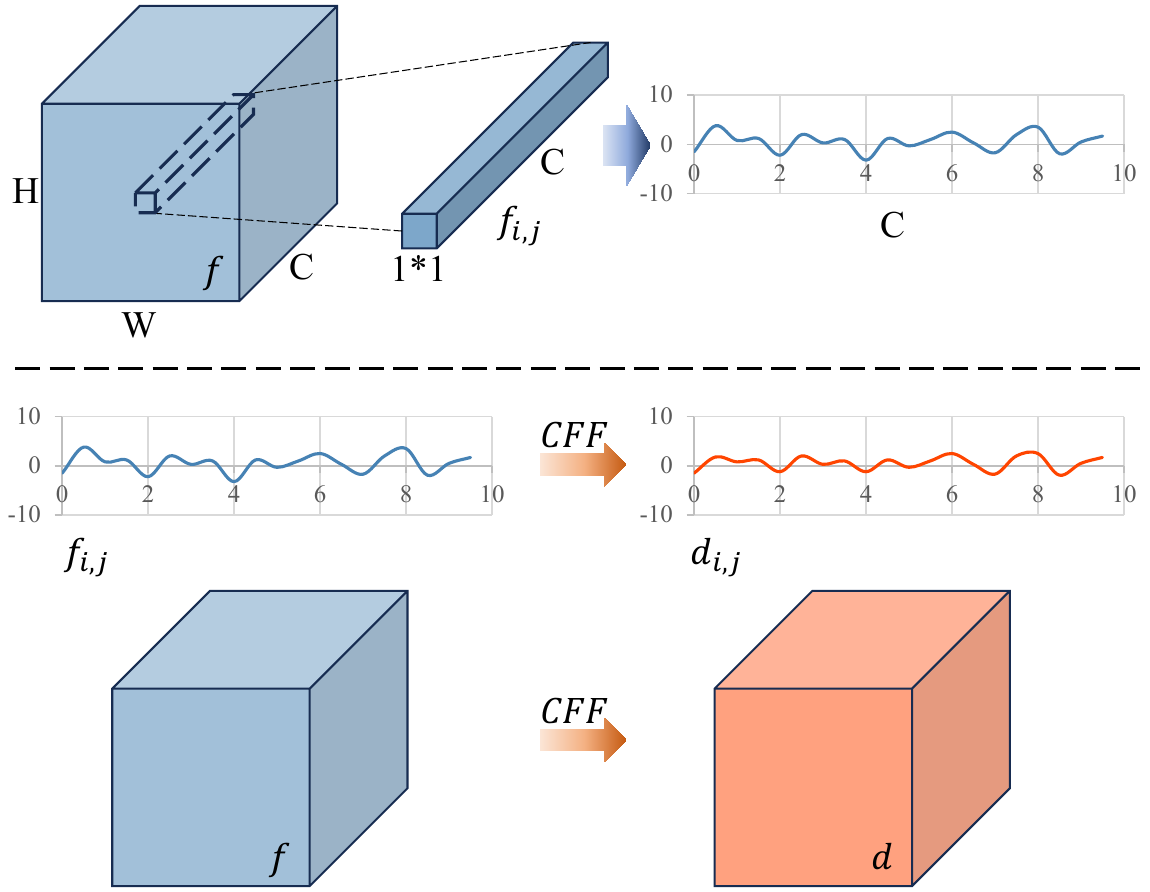}
\caption{An illustration for feature signal matrices, where $f$, $d$, H, W, and C correspond to a feature signal matrix, a filtered feature signal matrix, Height, Width, and Channel, respectively. The first row elucidates the rationale behind treating a 3D signal matrix as a 2D channel-based signal matrix, while the second row demonstrates the filtering process for a single-pixel signal and one signal matrix.}
\label{fig1}
\end{figure}

In this study, we consider convolution features following Gaussian distributions as \textbf{feature signals} in the feature space, intending to emulate the concept of frequency signals in the frequency space. The fundamental idea is that convolution operations are analogous to Fourier transformations, serving as non-invertible transformation functions. More simply and intuitively, a 3D feature signal matrix (ignoring batch size) with height, width, and channel of three dimensions can be equivalently regarded as a 2D signal matrix, and each pixel of it is a discrete channel-dimensional signal vector (depicted in the first row of Figure~\ref{fig1}). \textbf{Noise} is represented in the frequency space as a high-frequency form, while its representation takes the form of high amplitude in the feature space correspondingly. Based on the above conception and with the inspiration of the mechanism of DSP, we developed a feature filtering mechanism called the Convolutional Feature Filter (CFF) that aims to reduce noise in feature signal matrix inputs. It essentially is a \textbf{low-amplitude pass filter}, and the second row of Figure~\ref{fig1} illustrates an example of the filtering process for a single-pixel feature signal and one feature signal matrix. 

\begin{figure}[t]
\centering
\includegraphics[width=0.50\linewidth]{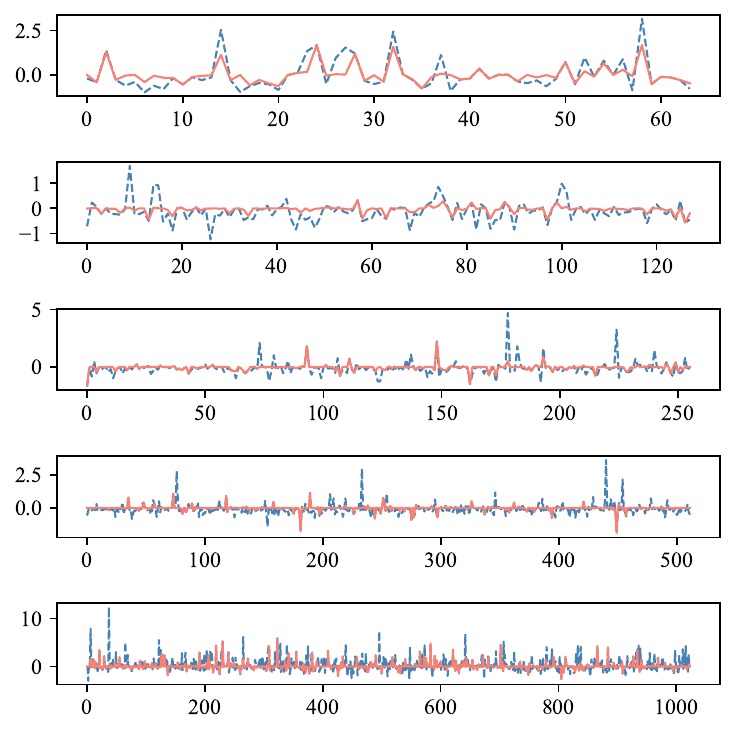}
\caption{Several examples of convolution feature signals with different channel numbers, where the x- and y-axes denote the channel number and amplitude, respectively. These feature signals are selected from the center of feature signal matrices when inputting a random CMRI to $Ne{u_u}$ for the single-domain cardiac segmentation task. The steel-blue dashed lines represent the original feature signals containing the high-peak noise. The orange-red lines depict the filtering results of those steel-blue lines, where the noise has been reduced.}
\label{fig2}
\end{figure}

We integrated the proposed CFF into each convolution layer of two established 2D segmentation networks, FCN~\cite{b1} and U-net~\cite{b3}, and these networks were evaluated on two public CMRI datasets and one public RGB image dataset. These datasets are renowned within the research community for their diverse and demanding characteristics, providing a robust benchmark for cardiac and semantic segmentation tasks. Two main objectives guide the experiments: one is dedicated to validating noise reduction, while the other focuses on observing the outcome of implementing noise reduction. The results of the first experimental objective demonstrated that the CFFs efficiently reduced noise in feature signal inputs, and Figure~\ref{fig2} exhibits several examples of feature signal processing. The orange-red lines depict the filtering outcomes, contrasting them with the original steel-blue dashed lines. As anticipated, the noise was successfully eliminated in the steel-blue dashed lines. In addition, we have identified an additional characteristic of noise in its probabilistic representation: its low probability. Therefore, we leveraged probability theory and employed information entropy~\cite{b9} as the evaluative metric for quantifying filtering changes. At last, the results of the second experimental objective showcased that the quality of the segmentation results has improved after incorporating CFFs. 

In conclusion, the \textbf{innovation} of this work is regarding convolution features as feature signal matrices and presenting corresponding operations. And based on it, we have two primary contributions as follows. 
(1) \textbf{An effective noise reduction in feature signals}. We put forth a convolutional feature filter. It being a high-probability and low-amplitude pass filter has advantages: simplicity, generality, and effectiveness. The first allows that incorporating it will not cause the burden of parameters for training a network, the second makes that it can be integrated into different segmentation networks for various segmentation tasks, and the last ensures that it can reduce the noise in feature signal inputs. 
(2) \textbf{A unified gate for subsequent noise reduction evaluation}. We develop a binarization equation for information entropy computation. First, it unifies probability matrices of feature signal metrics into the same scale, eliminating the influence of different sizes. Second, it divides the high- and low- probability signals based on the mean, resulting in two values more suitable for binary information entropy computation. Consequently, observing and analyzing noise reduction in the entire feature system is more straightforward.

\section{Methods}
\label{sec3}

\subsection{Convolutional Feature Filter}
\label{sec3a}
A convolutional feature filter (CFF) is designed with a simple structure exhibited as Formula (\ref{eq1}). Let $f \in {\mathbb R}_f^{H,W,ch}$ and $d \in {\mathbb R}_f^{H,W,ch}$ denote a given feature signal matrix and a filtered feature signal matrix, where ${{\mathbb R}_f}$ refers to the feature space and $H$, $W$, and $ch$ are their height, width, and channel number. The transformation of $f$ to $d$ can be described as
\begin{eqnarray}
d &=& cff(f,\theta )   \nonumber \\
~ &=& s(conv(f,\theta )) \cdot f,
\label{eq1}
\end{eqnarray}
where $conv( \cdot )$ refers to a convolution layer, and $\cdot $ means pixel-wise multiplication. $s( \cdot )$ is the activation function $sigmoid( \cdot )$, and $\theta$ represents trainable weights and biases of a convolution layer or a network and is not distinguished hereafter. The convolution kernel number of $conv( \cdot )$ is the same as the channel number of an input $f$, and the kernel size is $1 \times 1$. 

Considering one convolution layer is the kernel of a CFF due to the following reasons. First, including the existing low-pass filters, such as an ideal filter, a Butterworth filter, or a Gaussian filer, in a network convolution layer adds an extra hyperparameter. As the number of convolution layers increases, so does the number of hyperparameters that need confirmation through time-consuming experiments. Secondly, a complex inner structure may add numerous trainable weights and biases, increasing the training burden compared to the original network. Consequently, the optimal solution is to include a trainable convolution layer and naturally allow it to evolve into a filter with the anticipated function.

\subsection{Networks}
\label{sec3b}
This section introduces the four main networks. Firstly, there are two fundamental 2D segmentation networks known as FCN~\cite{b1} and U-net~\cite{b3}, which will be referred to as $Al{t_f}$ and $Al{t_u}$ hereinafter. $Al{t_f }$ and $Al{t_u}$ can be described as sets of convolutional, max-pooling, upsampling, and other operations. The main focus lies on their convolutional operations, with further elaboration provided in references~\cite{b2,b3}, albeit excluded from this discussion. A single convolutional operation typically consists of a convolutional layer and a batch normalization layer, and $Al{t_f}$ and $Al{t_u}$ each encompass $12$ and $22$ convolutional operations, respectively. We employ set $BC = \{ b{c_1},b{c_2},...,b{c_n}\}$, where $bc$ represents one convolutional operation and $n \in \{ 12,22\}$, for description. Then, we can elucidate the process of attaining a feature signal matrix $f$ by utilizing $bc$, which can be represented as
\begin{eqnarray}
f &=&bc(\alpha ,\theta ) \nonumber \\
~&=&n(r(conv(\alpha ,\theta ))),
\label{eq2}
\end{eqnarray}
where $\alpha$ refers to an input (a CMRI $x \in {{\mathbb R}^{H,W}}$ or $f$), $r( \cdot )$ is the activation function $relu( \cdot )$, and $n( \cdot )$ means a batch normalization. 

Appending the proposed CFFs into $Al{t_f}$ and $Al{t_u}$ obtained two new networks named $Ne{u_f}$ and $Ne{u_u}$, i.e., utilizing set $FBC = \{ fb{c_1},fb{c_2},...,fb{c_n}\}$, where $fbc( \cdot ) = cff(bc( \cdot ))$ is a combination of Formula (\ref{eq1}) and Formula (\ref{eq2}) and $n \in \{ 12,22\}$, to substitute set $BC$ in $Al{t_f}$ and $Al{t_u}$. Finally, we can succinctly depict the prediction process of the four primary networks as 
\begin{equation}
\hat y = \phi (x,\theta ),
\label{eq3}
\end{equation}
where $\phi $ repents $Al{t_f}$,  $Al{t_u}$,  $Ne{u_f}$, and $Ne{u_u}$ and $\hat y \in {{\mathbb R}^{H,W}}$ denotes the predicted label map.

\subsection{Entropy}
\label{sec3c}
\textbf{Global entropy}: We will begin by illustrating cross-entropy, a crucial function for leading the entire feature system of networks from disorder to order, and a lower validation entropy indicates a better network model. Here, a modified cross-entropy (CEA~\cite{b8}) is cited as the network training objective function, and it is
\begin{eqnarray}
L = {L_{CEA}}(y,\hat y), \nonumber \\
\mathop {\min }\limits_\theta  (y,L(\phi (x,\theta ))),
\label{eq4}
\end{eqnarray}
where $y \in {{\mathbb R}^{H,W}}$ denotes the ground-truth (GT) label map corresponding to an inputted $x$. Other details are in~\cite{b8}, and we do not elaborate too much. 

\textbf{Local entropy}: Introducing the information entropy~\cite{b9} plays a central role as a measure of information, choice, and uncertainty of a feature signal matrix. High-amplitude noise, constituting a minor portion of the feature signal matrix, exhibits a characteristic of low appearance frequency in its probabilistic representation, signifying a low probability. Thus, utilizing this characteristic to calculate the information entropy of feature signals will showcase the effectiveness of CFFs. Computing this entropy follows three steps: normalization, binarization, and computation. 

(\textbf{1}) $f$ and $d$ are subject to a normal distribution due to Formula (\ref{eq2}) and Theorem~\ref{th1}.
\begin{theorem}
\label{th1}
If $X$ is a $N(\mu ,{\sigma ^2})$ random variable and $Y = aX + b$, $a$ and $b$ are both real. Then $Y$ has a $N(a\mu + b,{a^2}{\sigma ^2})$ distribution~\cite{b13}.
\end{theorem}
The constant $\mu$, $\sigma $, and $\sigma ^2$ are the mean, standard deviation, and variance. In this work, $a$ is a filter matrix ($s(conv(f,\theta ))$ in Formula (\ref{eq1})) whose value is real, and $b$ is $0$. Regarding $f$ and $d$ as 1D random variables, we can utilize the density function of the normal distribution, 
\begin{equation}
p(\alpha ) = \frac{1}{{\sigma \sqrt {2\pi } }}{e^{\frac{{ - {{(\alpha  - \mu )}^2}}}{{2{\sigma ^2}}}}},\left| \alpha  \right| < \infty ,\left| \mu  \right| < \infty ,\sigma  > 0,
\label{eq5}
\end{equation}
$\alpha$ is an input, to normalize $f$ or $d$ and calculate the appearance frequency of signals for achieving a probability matrix, i.e., $pf = p(f)$, $ pd = p(d)$. 

(\textbf{2}) Binarizing $pf$ or $pd$ via
\begin{equation}
 P(\alpha ){\rm{ = }}\left\{ \begin{array}{l}
\{ 0.5, 0.5\} ,a \ge b\\
\{ a, b\} ,a < b
\end{array} \right.,a = \frac{{{n_\alpha }}}{{{N_\alpha }}},b = \frac{{{m_\alpha }}}{{{N_\alpha }}}, 
\label{eq6}
\end{equation}
i.e., $Pf = P(pf)$, $Pd = P(pd)$, is utilized to divide the high- and low-probability signals. $\alpha$ is an input, and ${{N_\alpha }}$, ${{n_\alpha }}$, and ${{m_\alpha }}$ are, respectively, the number of all elements, the number less than the mean, and the number more than the mean of an input. This designed formula unifies $pf$ or $pd$ with different sizes into the same scale, which benefits subsequent information entropy computing and comparing.

(\textbf{3}) The last step is computing the information entropy of $f$ or $d$ via
\begin{equation}
H(\alpha ) =  - K\sum {P(\alpha ) \cdot } {\log _2}(P(\alpha )),
\label{eq7}
\end{equation}
i.e., $Hf = H(Pf)$, $Hd = H(Pd)$. $\alpha$ is an input, and $K$ is a constant $1$. It is significant that normalization and binarization. Normalization first makes the mean of a feature signal matrix ($f$ or $d$) more valuable, and then binarization makes a probability matrix ($pf$ or $pd$) serve for computing the entropy of binary information. In addition, we can conclude the following information from Formula (\ref{eq6}) and Formula (\ref{eq7}). If $ a = 0.5$ and $b = 0.5$, or $a > b$, $H$ will be the maximum. If $a < b$ and $b \to 1$ and $a \to 0$, $H$ will be the minimum. 

Finally, given an input $x$ to $Ne{u_f}$ and $Ne{u_u}$, we can compute the information entropy of each feature signal matrix before and after filtering: $HF = \{ H{f_1}, H{f_2}, ..., H{f_n} \}$ to represent the results before filtering, and $HD = \{ H{d_1}, H{d_2}, ..., H{d_n} \}$ to represent the results after filtering. Then, we can compute the entropy variation to elucidate its physical significance, i.e., 
\begin{equation}
\Delta H = HD – HF.
\label{eq8}
\end{equation}
If $\Delta H < 0 $, CFFs work, feature signal matrices will evolve toward one with more information, better choice, and less uncertainty, and vice versa.

\section{Experiments}
\label{sec4}

\subsection{Data and Others}
\label{sec4a}
\textbf{Datasets}: Three distinct datasets were conducted in this study: a single-domain dataset of cardiac short-axis cine-MRI from the automated cardiac diagnosis challenge (ACDC)~\cite{b5}\footnote{ https://www.creatis.insa-lyon.fr/Challenge/acdc/ }, a multi-domain dataset of cardiac short-axis cine-MRI from the Multi-Centre, Multi-Vendor, and Multi-Disease (M\&Ms) challenge~\cite{b25}\footnote{https://www.ub.edu/mnms/}, and a single-domain RGB image dataset from the Pascal Visual Object Classes (VOC) Challenge~\cite{b15}\footnote{http://host.robots.ox.ac.uk/pascal/VOC/voc2012/}. The ACDC and M\&Ms datasets were utilized for the primary \textbf{cardiac segmentation} tasks, targeting the right ventricle (RV), myocardium (MYO), and left ventricle (LV). A subset (VOCH) of the VOC data containing humans served as the reference \textbf{object segmentation} task, focusing on segmentation targets such as human objects (HO) and borders or others (BO). 

\textbf{Data processing}: The ACDC dataset comprised three distinct subsets: a training set (TrC), a validation set (VaC), and a testing set (TeC). Similarly, the M\&Ms dataset was organized into three subsets: a training set (TrM), a validation set (VaM), and a testing set (TeM). As for the VOCH dataset, which involved extracting all human objects from the VOC data, it was divided into two subsets: a training set (TrV) and a validation set (VaV). We merged the end-diastolic (ED) and end-systolic (ES) phases of the training and validation data from the cardiac datasets while retaining the original phases for the testing data. Furthermore, we applied various image processing techniques, including cropping or padding size, data augmentation, and normalization. 

\textbf{Evaluation metrics}: We evaluated the segmentation accuracy using the {Dice} score and {Hausdorff distance (HD)} metrics. Higher Dice scores and lower HD distances indicate better performance.

\textbf{Networks}: In this study, we conducted ten different networks in our experimental environment. These included the main networks ($Al{t_f}$, $Al{t_u}$, $Ne{u_f}$, and $Ne{u_u}$), two ablation networks for studying the effects of specific components (CP1: ablation network on $Ne{u_f}$, CP2: ablation network on $Ne{u_u}$), and four comparison networks (M1: Transformer-based U-net, M2: CBAM-based U-net, CM1: M1 incorporating CFFs, and CM2: M2 incorporating CFFs).

\textbf{Implementation Details}: The entire set of experiments was performed on a single Nvidia RTX 4090 24 GB GPU, utilizing the Tensorflow-gpu 2.6.0 deep learning library. To ensure the effective validation of the proposed CFFs, we maintained consistent hyperparameters across all networks. These hyperparameters include 50 epochs, the Adam optimizer, a learning rate of 0.001, and a batch size of 1. Notably, each network underwent training at least ten times to guarantee robust and dependable results.

\textbf{Experiment Purposes}: Two main objectives guide the experiments: one is dedicated to validating noise reduction (see Section~\ref{sec4b} and Section~\ref{sec4c}), while the other focuses on observing the outcome of implementing noise reduction (see Section~\ref{sec4d}, Section~\ref{sec4e}, and Section~\ref{sec4f}).

\subsection{Global Entropy}
\label{sec4b}

\begin{figure}[t]
\centering
\begin{minipage}{0.75\linewidth}
     \begin{subfigure}[b]{0.45\linewidth}
     \includegraphics[width=\linewidth]{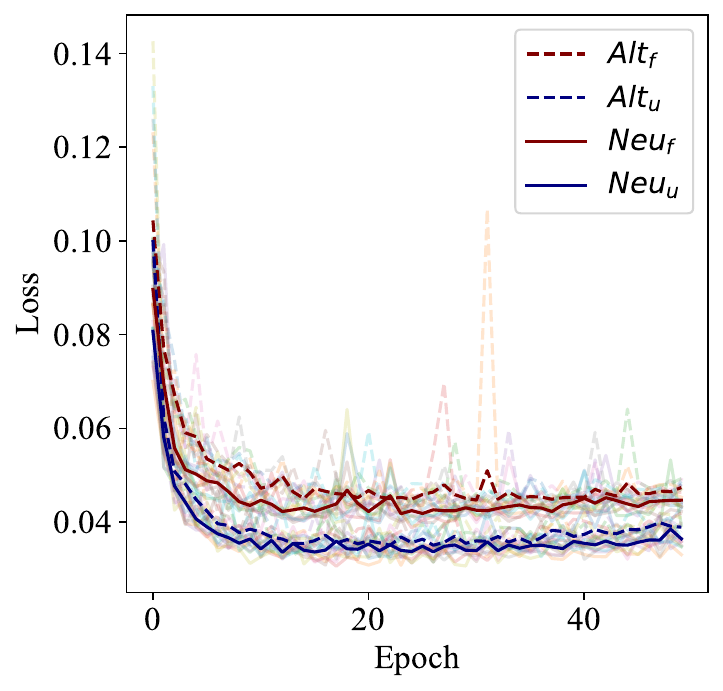}
     \caption{}
     \label{fig3a}
     \end{subfigure}
     \hfill
     \begin{subfigure}[b]{0.45\linewidth}
     \includegraphics[width=\linewidth]{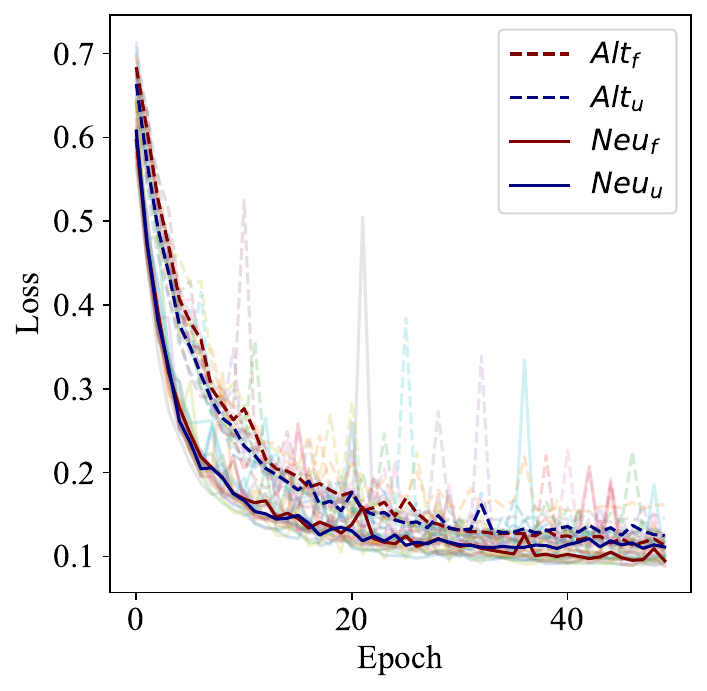}
     \caption{}
     \label{fig3b}
     \end{subfigure} 
\end{minipage}
\caption{Comparing validation results (validation loss) of global entropy across different segmentation tasks. Semi-translucent and opaque lines represent all results and their averages, respectively. The opaque lines with different colors indicate the different network results, whereas the opaque lines with the same color but dashed and full denote the results of the original network and this network after appending CFFs. (a) Single-domain Cardiac segmentation. (b) Object segmentation.}
\label{fig3}
\end{figure}

We collected the validation results of global entropy (validation loss) from the four main networks ($Al{t_f}$, $Al{t_u}$, $Ne{u_f}$, and $Ne{u_u}$) in different segmentation tasks and represented them in Figure~\ref{fig3} using semi-translucent lines. Figure~\ref{fig3a} displays the results for the single-domain cardiac segmentation task, and Figure~\ref{fig3b} presents the results for the object segmentation task. Moreover, we calculated the mean of those validation loss data and exhibited results as opaque lines in Figure~\ref{fig3} for easy observation.

Upon observing Figure~\ref{fig3a}, two results are concluded. Appending CFFs into networks leads to their more optimal solutions, and U-net performs better when applying data whose characteristics are similar to the ACDC data. Similarly, Figure~\ref{fig3b} provides two conclusions as well. CFFs are effective in non-medical segmentation tasks, and FCN demonstrating comparable or superior performance compared to U-net when applied to data resembling the characteristics of VOCH data. Finally, we conclude that the entire feature system of networks tends to evolve towards a state with lower uncertainty after appending CFFs, as evidenced by the observed reduction in validation global entropy.

\subsection{Information Entropy}
\label{sec4c}

\begin{figure}[t]
\centering
\begin{minipage}{0.9\linewidth}
     \begin{subfigure}[b]{0.50\linewidth}
     \includegraphics[width=1.0\linewidth]{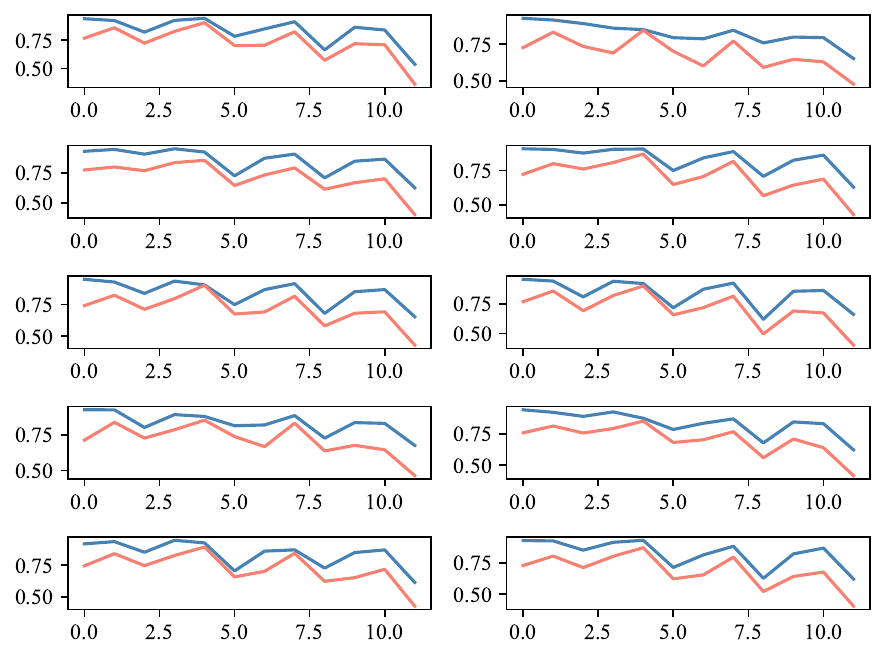}
     \caption{}
     \label{fig4a}
     \end{subfigure}
     \hfill
     \begin{subfigure}[b]{0.50\linewidth}
     \includegraphics[width=1.0\linewidth]{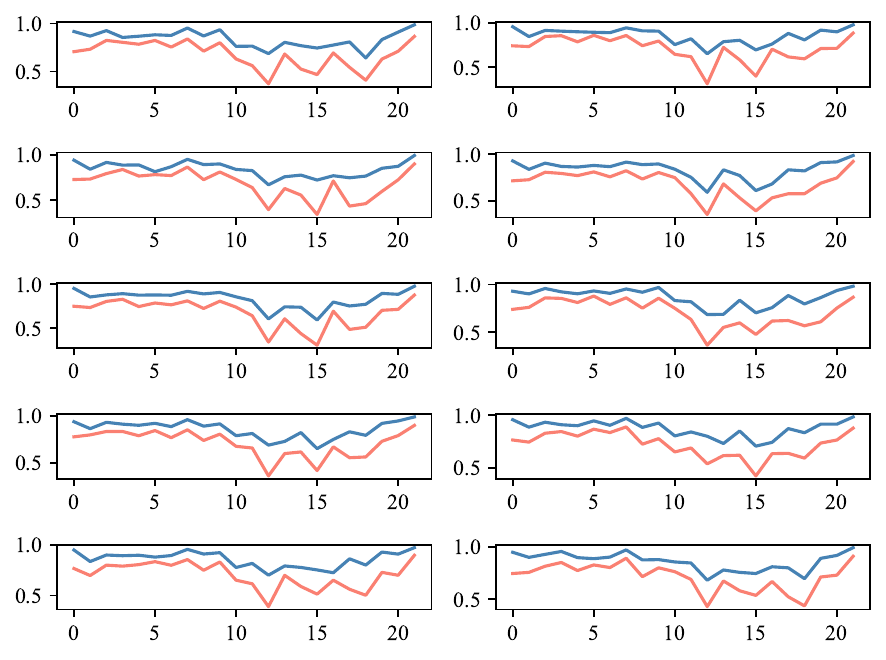}
     \caption{}
     \label{fig4b}
     \end{subfigure} 
\end{minipage}
\caption{Comparing the average information entropy on VaC across different networks. The steel-blue and orange-red lines represent $HF$ and $HD$, respectively. The x- and y-axes represent the number of elements within an information entropy set and the corresponding information entropy values, respectively. (a) Results of $Ne{u_f}$. (b) Results of $Ne{u_u}$.}
\label{fig4}
\end{figure}

We selected feature signal matrices of two networks, $Ne{u_f}$ and $Ne{u_u}$, to observe the information entropy variation after utilizing CFFs on the single-domain cardiac segmentation task. We fed VaC to each model of $Ne{u_f}$ and $Ne{u_u}$ and obtained the average information entropy sets: $HF$ and $HD$ for $Ne{u_f}$ and $Ne{u_u}$. Thus, we have ten results for each network, as illustrated in Figure~\ref{fig4}. Observing Figure~\ref{fig4a}, we can find two noticeable conclusions. The low-resolution feature signal matrices in a network have more information, and using CFFs for filtering has reduced the uncertainty of feature signal matrices. These findings are further reinforced through an analysis of Figure~\ref{fig4b}.

\begin{figure}[t]
\centering
\begin{minipage}{0.65\linewidth}
     \begin{subfigure}[b]{0.45\linewidth}
     \includegraphics[width=\linewidth]{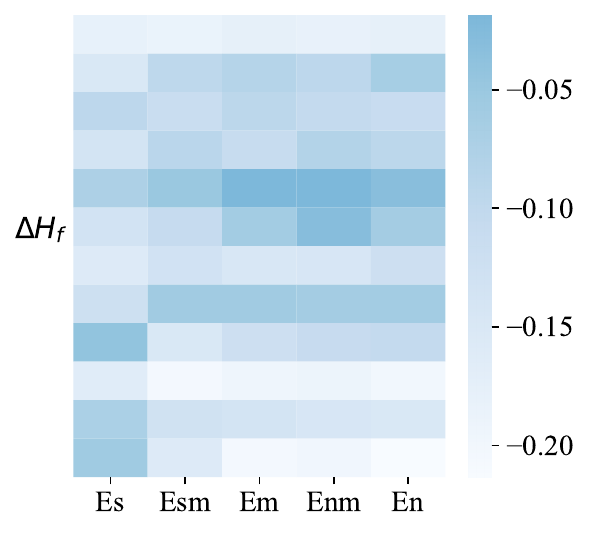}
     \caption{}
     \label{fig5a}
     \end{subfigure}
     \hfill
     \begin{subfigure}[b]{0.45\linewidth}
     \includegraphics[width=\linewidth]{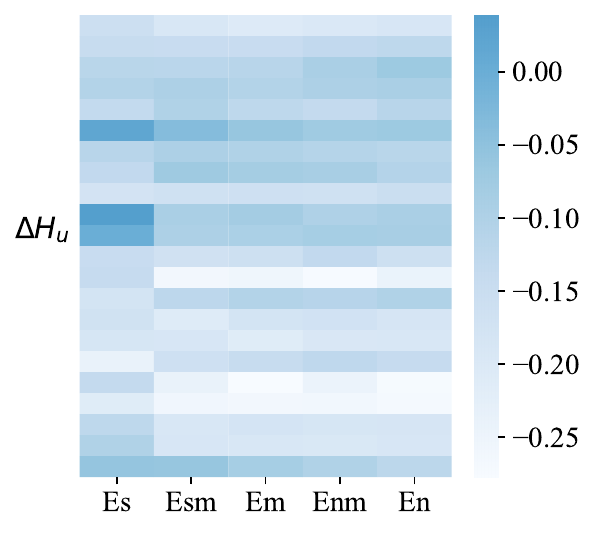}
     \caption{}
     \label{fig5b}
     \end{subfigure} 
\end{minipage}
\caption{Heat maps of the information entropy variation on VaC across different networks at different times, where the blue color goes light while the information entropy decrease, and vice versa. (a) $\Delta H$ for $Ne{u_f}$ ($\Delta {H_f}$). (b) $\Delta H$ for $Ne{u_u}$ ($\Delta {H_u}$).}
\label{fig5}
\end{figure}

In addition, observing $\Delta H$ (see Figure~\ref{fig5}) for one model of $Ne{u_f}$ and $Ne{u_u}$ at different times (Es: the first epoch, Esm: the middle epoch between En and Em, Em: the minimum validation loss epoch, Enm: the middle epoch between Em and En, and En: the last epoch), we conclude that the function of CFFs changed with changing times and that the optimal performance is during the minimum validation loss epoch. In conclusion, both Figure~\ref{fig4} and Figure~\ref{fig5} verified that the proposed CFF effectively reduces noise and that the validation loss decrease is related to the declined local entropy of feature signals.

\subsection{Ablation Study and ACDC Segmentation}
\label{sec4d}

First, let's present the ablation validation of the kernel size ($1 \times 1$) for a CFF, and this experiment was performed on the single-domain cardiac segmentation task. Supposing that $Al{t_f}$ and $Al{t_u}$ include $k$ parameters, we appended CFFs to them, then $Ne{u_f}$ and $Ne{u_u}$ have $1.1k$ parameters. However, by changing the kernel size of CFFs to $3 \times 3$, the FCN and U-net would contain $2.02k$ parameters, and we named them CP1 and CP2. Table~\ref{tab3} presents the segmentation evaluations on VaC, showing the mean and standard deviation (Std) values obtained from $Al{t_f}$, $Al{t_u}$, $Ne{u_f}$, $Ne{u_u}$, CP1, and CP2. The results demonstrate that, for single-domain cardiac segmentation, the U-Net architecture outperforms the FCN. Moreover, networks that incorporate CFFs exhibit better validation results, and the current kernel size ($1 \times 1$) of CFFs is a more optimal selection. 

\begin{table}[t]
\centering
\caption{Evaluation metrics on VaC (Mean $\pm$ Std). Mean Seg. denotes the average results of LV, RV, and MYO segmentation. Bold highlights indicate improvements, while red indicates the best results.}
\begin{tabular}{lcc}
\hline
Mean Seg.&Dice (\%)&HD (mm)\\
\hline
$Al{t_f}$ & 84.5 $\pm$ 0.7 & 10.2 $\pm$ 0.7 \\
CP1 & \textbf{85.1 $\pm$ 0.8} & \textbf{9.1 $\pm$ 0.8}\\ 
$Ne{u_f}$ &\textbf{85.3 $\pm$ 0.6} & \textbf{9.3 $\pm$ 0.6} \\ 
\hline
$Al{t_u}$ & 86.9 $\pm$ 0.6 & 7.7 $\pm$ 0.6 \\ 
CP2 &\textbf{87.0 $\pm$ 0.8} &\textbf{7.3 $\pm$ 0.7} \\ 
$Ne{u_u}$ &\textcolor{red}{\textbf{87.4 $\pm$ 0.4}} & \textcolor{red}{\textbf{7.1 $\pm$ 0.5}} \\
\hline
\end{tabular}
\label{tab3}
\end{table}

\begin{table}[t]
\centering
\caption{Evaluation metrics on TeC (Mean $\pm$ Std, Dice (\%), HD (mm)). Bold highlights indicate improvements, while red indicates the best results.}
\begin{tabular}{lcccc}
\hline
LV&Dice ED&Dice ES&HD ED&HD ES\\
\hline
$Al{t_f}$&96.0$\pm$0.3&90.6$\pm$0.6&8.6$\pm$1.6&17.1$\pm$2.8\\ 
$Ne{u_f}$&\textbf{96.1$\pm$0.2}&\textbf{90.9$\pm$0.3}&\textbf{7.2$\pm$1.4}&\textbf{14.2$\pm$2.1}\\ 
\hline
$Al{t_u}$&96.4$\pm$0.2&91.7$\pm$0.7&4.7$\pm$0.4&5.3$\pm$1.3\\ 
$Ne{u_u}$&\textcolor{red}{\textbf{96.6$\pm$0.1}}&\textcolor{red}{\textbf{92.3$\pm$0.5}}&\textcolor{red}{\textbf{3.6$\pm$0.4}}&\textcolor{red}{\textbf{4.1$\pm$0.6}}\\
\hline
\hline
RV&Dice ED&Dice ES&HD ED&HD ES\\
\hline
$Al{t_f}$&92.8$\pm$0.4&86.5$\pm$0.8&16.2$\pm$2.8&19.4$\pm$ 3.6\\ 
$Ne{u_f}$&\textbf{92.9$\pm$0.3}&\textbf{86.5$\pm$0.6}&\textbf{15.1$\pm$1.8}&\textbf{16.5$\pm$1.7}\\ 
\hline
$Al{t_u}$&93.4$\pm$0.5&87.6$\pm$1.1&12.0$\pm$2.1&11.6$\pm$2.2\\ 
$Ne{u_u}$&\textcolor{red}{\textbf{93.8$\pm$0.4}}&\textcolor{red}{\textbf{88.2$\pm$0.9}}&\textcolor{red}{\textbf{11.6$\pm$2.3}}&\textbf{\textcolor{red}{11.2$\pm$1.2}}\\
\hline
\hline
MYO&Dice ED&Dice ES&HD ED&HD ES\\
\hline
$Al{t_f}$&87.6$\pm$0.3&89.3$\pm$0.3&12.3$\pm$2.2&21.5$\pm$2.5\\ 
$Ne{u_f}$&\textbf{87.8$\pm$0.3}&\textbf{89.3$\pm$0.2}&\textbf{10.4$\pm$ 2.5}&\textbf{16.4$\pm$2.1}\\ 
\hline
$Al{t_u}$&89.3$\pm$0.3&90.9$\pm$0.3&5.7$\pm$1.1&5.9$\pm$1.5\\ 
$Ne{u_u}$&\textcolor{red}{\textbf{89.7$\pm$0.2}}&\textbf{\textcolor{red}{91.2$\pm$0.2}}&\textcolor{red}{\textbf{5.1$\pm$0.6}}&\textbf{\textcolor{red}{5.6$\pm$1.2}}\\
\hline
\end{tabular}
\label{tab4}
\end{table}

Then, we observe the segmentation evaluations on TeC (depicted in Table~\ref{tab4}), showing the mean and Std values obtained from $Al{t_f}$, $Al{t_u}$, $Ne{u_f}$, and $Ne{u_u}$. The results demonstrate that networks incorporating CFFs exhibit improved segmentation evaluations, consistent with the findings in Table~\ref{tab3}. Notably, the best evaluation always belongs to networks appended CFFs. Thus, it verified that noise reduction has benefits for improving segmentation accuracy. 

\subsection{M\&Ms Segmentation}
\label{sec4e}

This experiment aims to demonstrate the performance of CFFs when the networks handle the scenario that the data are from multi-domain. In order to better showcase the benefits of CFFs, we additionally integrated two widely recognized and popular attention mechanisms, namely Transformer and CBAM, into the U-net backbone. Thus, this experiment comprises a total of eight networks: the four main networks and M1, CM1, M2, and CM2, and the segmentation evaluations on VaM and TeM are shown in Table~\ref{tab5}, Table~\ref{tab6}, and Table~\ref{tab7}.

\begin{table}[t]
\centering
\caption{Evaluation metrics on VaM (Mean $\pm$ Std). Mean Seg. denotes the average results of LV, RV, and MYO segmentation. Bold highlights indicate improvements, while red indicates the best results.}
\begin{tabular}{lcc}
\hline
Mean Seg.&Dice (\%)&HD (mm)\\
\hline
$Al{t_f}$ & 80.1 $\pm$ 2.8 & 22.3 $\pm$ 5.9 \\
$Ne{u_f}$ & \textbf{83.0 $\pm$ 0.8} & \textbf{17.8 $\pm$ 1.2} \\
\hline
$Al{t_u}$ & 82.5 $\pm$ 1.4 & 17.7 $\pm$ 2.4 \\
$Ne{u_u}$ & \textbf{84.9 $\pm$ 1.3} & \textbf{14.4 $\pm$ 1.9} \\
\hline 
M1 & 82.9 $\pm$ 0.9 & 17.5 $\pm$ 2.0 \\
CM1 & \textbf{84.1 $\pm$ 1.2} & \textbf{15.9 $\pm$ 1.8} \\
\hline
M2 & 85.0 $\pm$ 2.8 & 17.9 $\pm$ 3.9 \\
CM2 & \textcolor{red}{\textbf{85.2 $\pm$ 0.8}} &  \textcolor{red}{\textbf{14.3 $\pm$ 1.2}} \\
\hline
\end{tabular}
\label{tab5}
\end{table}

\begin{table}[t]
\centering
\caption{Evaluation metrics on TeM (Mean $\pm$ Std, Dice (\%), HD (mm)). Bold highlights indicate improvements, while red indicates the best results.}
\begin{tabular}{lcccc}
\hline
LV&Dice ED&Dice ES&HD ED&HD ES\\
\hline
$Al{t_f}$&79.8$\pm$2.2&73.4$\pm$1.9&28.5$\pm$11.7&29.8$\pm$12.0\\
$Ne{u_f}$&\textcolor{red}{\textbf{80.7$\pm$1.7}}&\textbf{74.8$\pm$1.2}&\textbf{20.2$\pm$2.0}&\textbf{22.5$\pm$3.3}\\
\hline
$Al{t_u}$&76.9$\pm$2.3&73.3$\pm$2.2&24.6$\pm$6.7&26.3$\pm$5.9\\
$Ne{u_u}$&\textbf{78.2$\pm$2.9}&\textcolor{red}{\textbf{76.6$\pm$1.4}}&\textbf{20.8$\pm$3.5}&\textcolor{red}{\textbf{18.8$\pm$2.0}}\\
\hline
M1&75.6$\pm$1.8&72.9$\pm$1.5&22.9$\pm$4.1&24.4$\pm$4.0\\
CM1&\textbf{77.3$\pm$2.1}&\textbf{73.9$\pm$1.4}&\textbf{21.0$\pm$3.3}&\textbf{20.6$\pm$2.8}\\
\hline
M2&77.5$\pm$3.1&74.5$\pm$2.4&20.3$\pm$2.9&19.9$\pm$2.4\\
CM2&\textbf{80.0$\pm$1.5}&\textbf{76.6$\pm$1.6}&\textcolor{red}{\textbf{19.8$\pm$2.5}}&\textbf{19.0$\pm$2.3}\\
\hline
\hline
RV&Dice ED&Dice ES&HD ED&HD ES\\
\hline
$Al{t_f}$&87.5$\pm$1.7&80.3$\pm$1.2&24.2$\pm$6.2&28.3$\pm$8.2\\
$Ne{u_f}$&\textbf{88.7$\pm$1.4}&\textbf{82.2$\pm$1.2}&\textbf{17.9$\pm$4.3}&\textbf{20.5$\pm$4.3}\\
\hline
$Al{t_u}$&87.4$\pm$1.4&81.1$\pm$1.0&16.3$\pm$5.3&21.8$\pm$6.7\\
$Ne{u_u}$&\textbf{88.5$\pm$1.5}&\textbf{83.6$\pm$0.7}&\textcolor{red}{\textbf{9.4$\pm$1.9}}&\textcolor{red}{\textbf{12.1$\pm$2.2}}\\
\hline
M1&86.3$\pm$1.7&81.3$\pm$1.1&17.4$\pm$5.0&21.8$\pm$6.5\\
CM1&\textbf{86.6$\pm$2.1}&\textbf{81.8$\pm$1.7}&\textbf{15.4$\pm$3.0}&\textbf{20.2$\pm$4.3}\\
\hline
M2&88.3$\pm$1.6&82.2$\pm$1.5&14.6$\pm$5.2&18.5$\pm$6.4\\
CM2&\textcolor{red}{\textbf{89.1$\pm$1.0}}&\textcolor{red}{\textbf{84.3$\pm$0.6}}&\textbf{12.4$\pm$2.6}&\textbf{15.4$\pm$2.6}\\
\hline
\hline
MYO&Dice ED&Dice ES&HD ED&HD ES\\
\hline
$Al{t_f}$&75.2$\pm$1.2&76.8$\pm$1.1&28.2$\pm$7.0&32.0$\pm$8.5\\
$Ne{u_f}$&\textbf{76.6$\pm$1.2}&\textbf{78.4$\pm$1.2}&\textbf{22.3$\pm$3.4}&\textbf{26.4$\pm$3.7}\\
\hline
$Al{t_u}$&75.2$\pm$1.3&74.4$\pm$1.4&20.0$\pm$6.0&23.5$\pm$7.7\\
$Ne{u_u}$&\textbf{77.2$\pm$1.2}&{\textbf{80.8$\pm$0.7}}&\textcolor{red}{\textbf{12.5$\pm$2.4}}&\textcolor{red}{\textbf{14.5$\pm$3.3}}\\
\hline
M1&75.4$\pm$1.7&77.8$\pm$1.4&23.2$\pm$8.1&26.9$\pm$9.3\\
CM1&\textbf{76.0$\pm$1.4}&\textbf{78.8$\pm$1.7}&\textbf{19.5$\pm$4.8}&\textbf{23.5$\pm$4.3}\\
\hline
M2&75.8$\pm$2.2&78.9$\pm$1.3&15.6$\pm$4.3&17.6$\pm$5.0\\
CM2&\textcolor{red}{\textbf{78.1$\pm$0.8}}&\textcolor{red}{\textbf{81.0$\pm$0.9}}&\textbf{13.7$\pm$1.9}&\textbf{16.5$\pm$2.6}\\
\hline
\end{tabular}
\label{tab6}
\end{table}

\begin{table*}[t]
\centering
\caption{Mean Dice metrics in different domains on TeM. Average Loss for Domain D: $average (D - ((A+B+C)/3))$. Bold highlights indicate improvements, while red indicates the best results.}
\begin{tabular}{lccccccccccccc}
\hline
Dice&\multicolumn{3}{c}{{Domain A}}&\multicolumn{3}{c}{{Domain B}}&\multicolumn{3}{c}{{Domain C}}&\multicolumn{3}{c}{{Domain D}}&\multicolumn{1}{c}{{ Loss for D}} \\
(\%) &\multicolumn{1}{c}{{LV}}&\multicolumn{1}{c}{{RV}}&\multicolumn{1}{c}{{MYO}}& \multicolumn{1}{c}{{LV}}& \multicolumn{1}{c}{{RV}}&\multicolumn{1}{c}{{MYO}}& \multicolumn{1}{c}{{LV}}& \multicolumn{1}{c}{{RV}}&\multicolumn{1}{c}{{MYO}}& \multicolumn{1}{c}{{LV}}& \multicolumn{1}{c}{{RV}}&\multicolumn{1}{c}{{MYO}}& \multicolumn{1}{c}{{Average}} \\
\hline
$Al{t_f}$&85.1&91.1&79.5&85.3&88.5&82.4&74.2&83.5&75.2&66.9&76.8&68.9&11.9\\
$Ne{u_f}$&\textbf{86.0}&\textbf{91.9}&\textbf{80.3}&\textbf{86.3}&\textbf{89.1}&\textbf{82.8}&\textcolor{red}{\textbf{74.9}}&\textbf{85.3}&\textbf{76.8}&\textbf{68.7}&\textbf{79.3}&\textbf{71.8}&\textbf{10.5}\\
\hline
$Al{t_u}$&85.7&91.1&80.7&85.5&88.9&83.5&71.5&82.2&74.5&64.1&78.7&69.2&12.0\\
$Ne{u_u}$&\textcolor{red}{\textbf{87.6}}&\textbf{92.3}&\textbf{82.5}&\textcolor{red}{\textbf{87.3}}&\textbf{89.8}&\textbf{85.3}&\textbf{73.8}&\textbf{84.3}&\textbf{77.0}&\textbf{66.9}&\textbf{81.5}&\textbf{73.5}&\textbf{10.5}\\
\hline
M1&86.9&92.3&82.2&86.5&89.8&84.5&71.3&81.8&74.6&60.0&76.4&68.5&15.1\\
CM1&86.9&92.1&82.2&\textbf{86.7}&\textbf{90.0}&\textbf{85.0}&\textbf{71.6}&81.8&\textbf{74.9}&\textbf{63.8}&\textbf{77.7}&\textbf{70.5}&\textbf{12.8}\\
\hline
M2&86.4&91.1&81.1&85.6&89.1&84.0&73.6&84.0&76.1&64.6&80.3&70.6&11.6\\
CM2&\textbf{86.8}&\textcolor{red}{\textbf{93.0}}&\textcolor{red}{\textbf{82.8}}&\textbf{86.9}&\textcolor{red}{\textbf{90.1}}&\textcolor{red}{\textbf{85.3}}&\textbf{74.8}&\textcolor{red}{\textbf{85.5}}&\textcolor{red}{\textbf{77.8}}&\textcolor{red}{\textbf{69.7}}&\textcolor{red}{\textbf{81.9}}&\textcolor{red}{\textbf{74.1}}&\textcolor{red}{\textbf{9.5}}\\
\hline
\end{tabular}
\label{tab7}
\end{table*}

Table~\ref{tab5} and Table~\ref{tab6} present the evaluation results of all networks on VaM and TeM. These results strongly support the idea that incorporating CFFs leads to a reduction in feature signal noise that causes an increase in Dice metrics and a decrease in HD metrics. However, the metrics in these tables are lower than those shown in Table~\ref{tab3} and Table~\ref{tab4}, which highlights the limitations of networks when dealing with unseen distribution data. In multi-domain cardiac segmentation, the crucial evaluation revolves around the loss incurred when dealing with the unseen domain D compared to known domains A, B, and C, as depicted in Table~\ref{tab7}. The results indicate that incorporating CFFs can improve the unseen domain segmentation accuracy. The losses for the unseen domain emphasize the positive impact of our proposed filter mechanism on the generalization ability of networks, making our work highly meaningful.

\subsection{VOCH Segmentation}
\label{sec4f}

\begin{table}[t]
\centering
\caption{Evaluation metrics on VaV (Mean $\pm$ Std). Bold highlights indicate improvements, while red indicates the best results.}
\begin{tabular}{lcc}
\hline
HO&Dice (\%)&HD (mm)\\
\hline
$Al{t_f}$ & 95.6 $\pm$ 0.2 & 46.0 $\pm$ 4.0 \\
$Ne{u_f}$ & \textcolor{red}{\textbf{96.4 $\pm$ 0.1}}&  \textcolor{red}{\textbf{34.9 $\pm$ 4.2}}\\
\hline
$Al{t_u}$ & 95.3 $\pm$ 0.6 & 48.9 $\pm$ 5.0 \\ 
$Ne{u_u}$ & \textbf{96.0 $\pm$ 0.4} &  \textbf{42.8 $\pm$ 5.1} \\
\hline
\hline
BO&Dice (\%)&HD (mm)\\
\hline
$Al{t_f}$ & 66.7 $\pm$ 1.2 & 59.7 $\pm$ 3.0 \\
$Ne{u_f}$ & \textbf{71.6 $\pm$ 1.0} & \textcolor{red}{\textbf{48.0 $\pm$ 3.2}} \\
\hline
$Al{t_u}$ & 69.1 $\pm$ 2.4 & 70.8 $\pm$ 6.9 \\ 
$Ne{u_u}$ &  \textcolor{red}{\textbf{72.6 $\pm$ 1.7}} & \textbf{64.9 $\pm$ 5.1} \\
\hline
\end{tabular}
\label{tab8}
\end{table}

This experiment aims to demonstrate the performance of CFFs when the designed networks process the non-medical data. Given the self-built dataset and the reference segmentation task, we focused solely on implementing the four main networks, and Table~\ref{tab8} demonstrates the segmentation evaluations on VaV. The results indicate that all the better and even the best metrics belong to the networks that incorporated CFFs. This finding leads to the conclusion that CFFs are well performed even in non-medical image segmentation tasks and proves the generality of CFFs. Furthermore, we noted that the FCN performs as well as or better than the U-net, consistent with the findings in Section~\ref{sec4b}. It underscores the task-specific adaptability and efficacy of networks.

\section{Related Works}
\label{sec2}

\textbf{Segmentation Networks}: The introduction of the Fully Convolutional Network (FCN) for semantic segmentation by J. Long et al.~\cite{b1}  in 2015 sparked significant interest among researchers in employing deep learning techniques for medical image segmentation, leading to notable contributions from various works, including those by P.V. Tran~\cite{b2} , M. Khened et al.~\cite{b16}, and others. One popular architecture that effectively incorporates context information into higher resolution layers is the U-net~\cite{b3} , proposed by O. Ronneberger et al., which has become a widely used backbone in segmentation studies, such as 3D U-net~\cite{b24}, ACNNs~\cite{b6}, Cross U-net ~\cite{b8},  U-net incorporating shape priors~\cite{b32}, and nnU-net~\cite{b28}. 

\textbf{Attention Mechanisms}: Attention Mechanisms (AMs), such as the renowned transformer~\cite{b10}, CBAM~\cite{b11}, and Attention-Conditioned Augmentation~\cite{b34}, have achieved significant success in medical segmentation studies. They mimic the human visual system, enabling them to focus on relevant parts in the inputs. 

\section{Conclusion}
\label{sec5}
In this work, we brought up a notion that regards convolution features as feature signal matrices. Based on it, a low-amplitude pass filter named CFF is designed for noise reduction in feature signal inputs. In addition, we developed Formula (\ref{eq6}) served as a unified gate before information entropy computing and comparing. Despite achieving the demand function, numerous limitations of the CFF persist. For example, its validation is lacking across other distribution feature signals, other research datasets, classification tasks, and even semi-supervised learning approaches. Thus, our forthcoming efforts will focus on addressing these limitations. Ultimately, we hold the belief that CFFs will play a pivotal and invaluable role in deep learning-based methodologies. Furthermore, we aspire for information entropy to serve as a standardized metric for evaluating feature fusion, selection, concatenation, and analysis. 

{
\bibliographystyle{IEEEbib}
\bibliography{icme2025references}
}

\newpage
\appendix

\subsection{Experimental Details: Images and Tables}

\begin{figure}[h]
\centering
\includegraphics[width=0.45\linewidth]{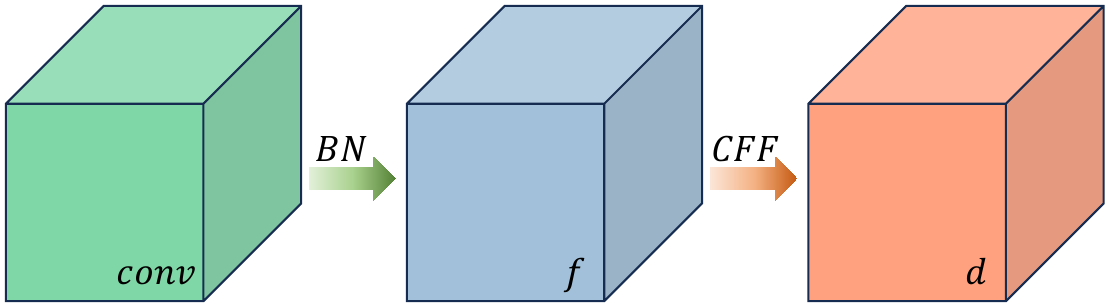}
\caption{A transformation from $f$ to $d$, where $conv$ and $BN$ indicate the original feature matrix and Batch normalization operation.}
\label{fig6}
\end{figure}

\begin{figure}[h]
\centering
\begin{minipage}{0.65\linewidth}
     \begin{subfigure}[b]{0.49\linewidth}
     \includegraphics[width=0.99\linewidth]{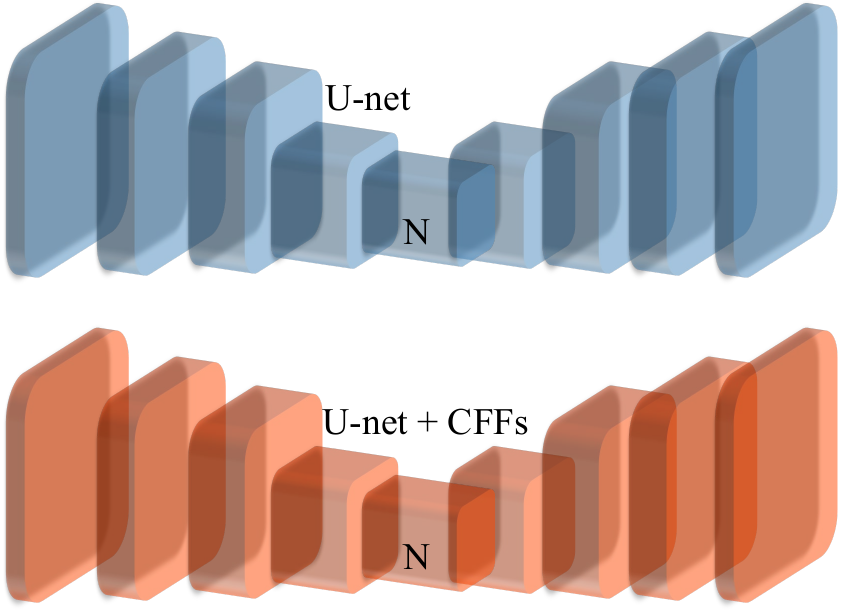}
     \caption{}
     \label{fig7a}
     \end{subfigure}
     \hfill
     \begin{subfigure}[b]{0.49\linewidth}
     \includegraphics[width=0.79\linewidth]{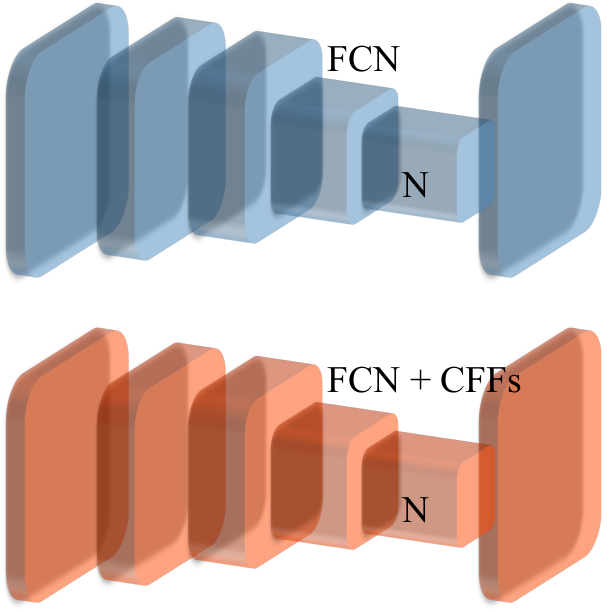}
     \caption{}
     \label{fig7b}
     \end{subfigure} 
     \vfill
     \begin{subfigure}[b]{0.49\linewidth}
     \includegraphics[width=0.99\linewidth]{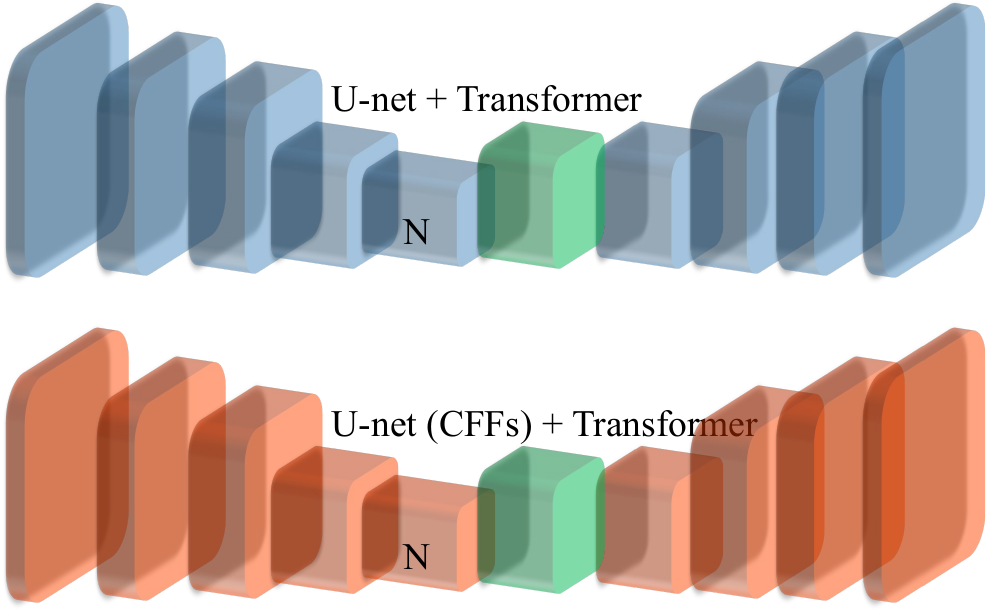}
     \caption{}
     \label{fig7c}
     \end{subfigure}
     \hfill
     \begin{subfigure}[b]{0.49\linewidth}
     \includegraphics[width=0.99\linewidth]{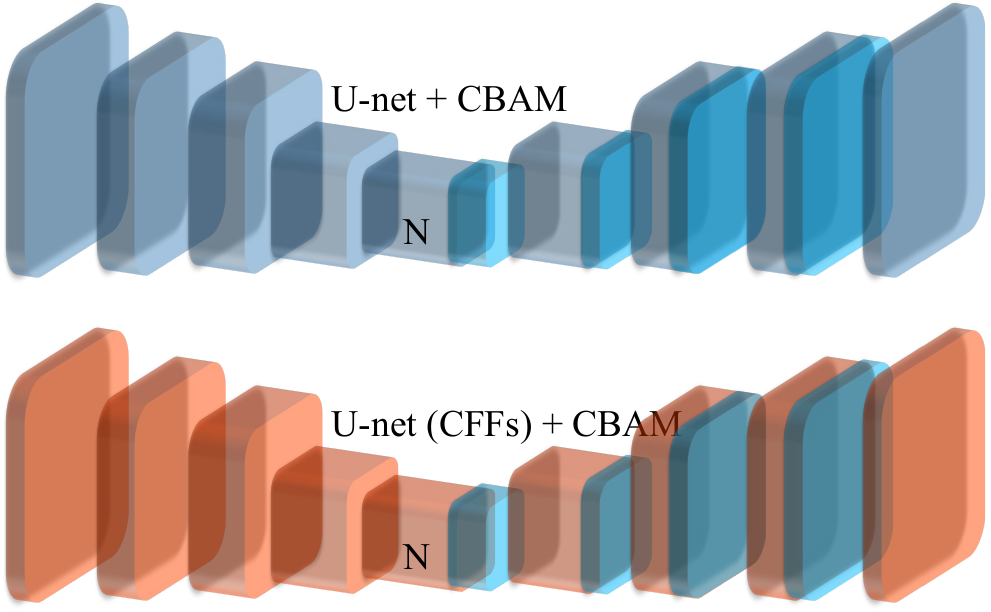}
     \caption{}
     \label{fig7d}
     \end{subfigure} 
\end{minipage}
\caption{Network Structures. Steel-blue, orange-red, light-green, and sky-blue blocks denote Convolutional layers with Batch normalizations, Convolutional layers with Batch normalizations and CFFs, transformer blocks, and CBAMs. N: the maximum channel number (1024). (a) $Al{t_u}$ and $Ne{u_u}$. (b) $Al{t_f}$ and $Ne{u_f}$. (c) M1 and CM1. (d) M2 and CM2.}
\label{fig7}
\end{figure}

The transformation from $f$ to $d$ is illustrated in Figure~\ref{fig6}. The network structures are presented in Figure~\ref{fig7}. Detailed information on the processed data is provided in Table~\ref{tab1}, while Table~\ref{tab2} displays the network parameters.

\begin{table*}[h]
\centering
\caption{Data information. Tr., Va., and Te. indicate training, validation, and testing. Seg. denotes Segmentation. NM: Normalization, RR: Random Rotation, RT: Random Translation, GB: Gaussian Blur, GN: Gaussian Noise.}
\begin{tabular}{lccccc}
\hline
Dataset&Domain&Seg. Targets&Size&Data Augmentation&Number of Samples\\
{}&{}&{}&{}&{}&Tr.(image)/Va.(image)/Te.(case)\\
\hline
ACDC& Single & LV/RV/MYO & $160^2$ & NM / RR & TrC: 6064 / VaC: 1544 / TeC: 50\\
M\&Ms& Multiple & LV/RV/MYO& $256^2$ & NM / RT / GB & TrM: 11580 / VaM: 2418 / TeM: 136\\
VOCH& Single & HO/BO & $512^2$ & NM / RT / GB / GN & TrV: 6776 / VaV: 968 / - \\
\hline
\end{tabular}
\label{tab1}
\end{table*}

\begin{table}[h]
\centering
\caption{Parameters of Networks. M: million.}
\begin{tabular}{cc}
\hline
Network & Parameter (M)\\
\hline
$Al{t_f}$ / $Ne{u_f}$ / CP1 & 31.30 / 35.48 / 68.91 \\
$Al{t_u}$ / $Ne{u_u}$ / CP2 & 34.53 / 38.38 / 69.08\\
M1 / CM1 & 71.28 / 75.12 \\
M2 / CM2 & 34.72 / 38.56\\
\hline
\end{tabular}
\label{tab2}
\end{table}

\begin{figure}[h]
\centering
\includegraphics[width=0.9\linewidth]{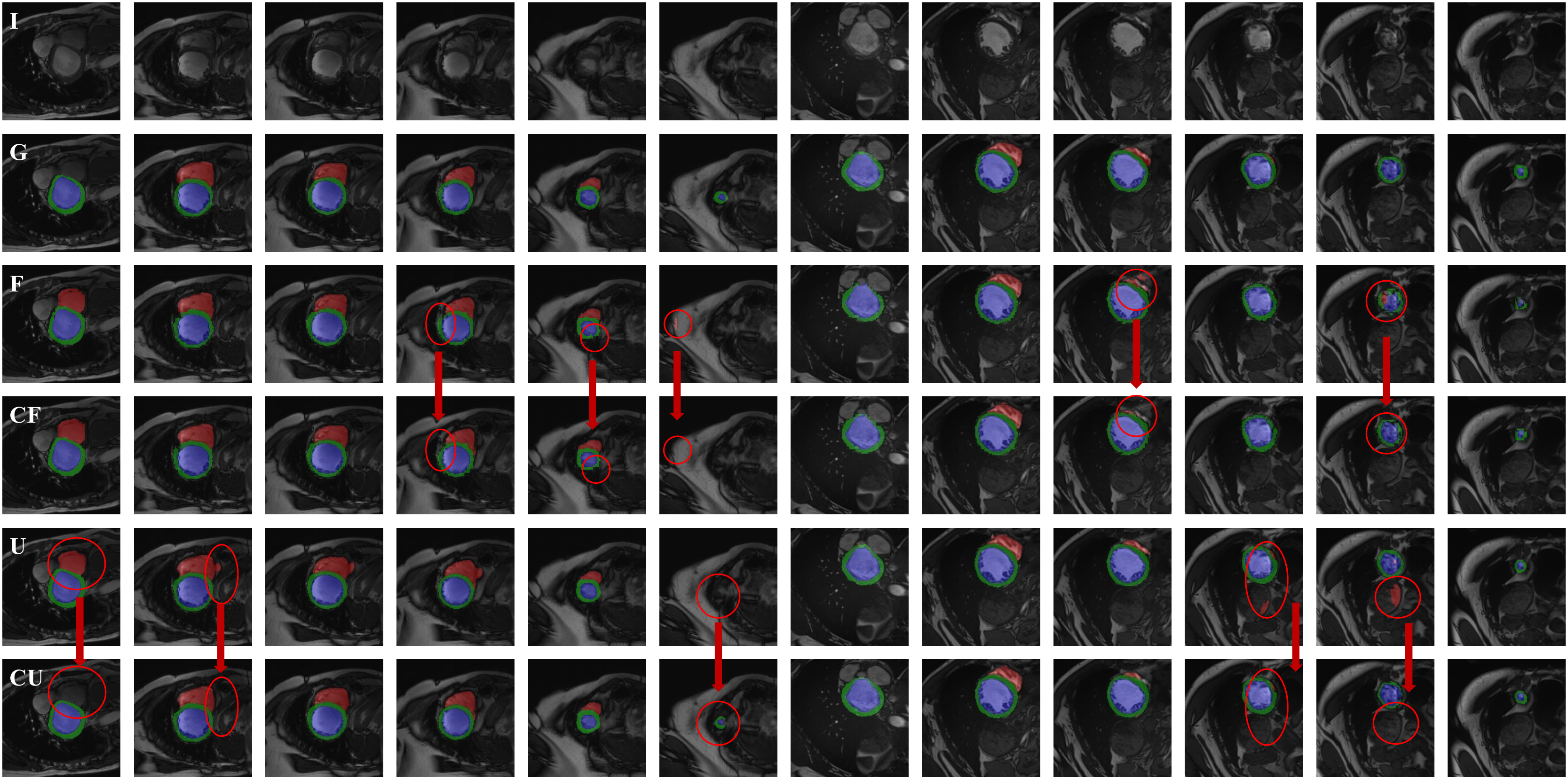}
\caption{Segmentation results on VaC obtained from different networks. The I and G are images and corresponding GT label maps, the U and CU are the results obtained from $Al{t_u}$ and $Ne{u_u}$, and the F and CF are the results obtained from $Al{t_f}$ and $Ne{u_f}$. The blue, green, and red parts are LV, MYO, and RV, respectively. }
\label{fig8a}
\end{figure}

Noticing Figure~\ref{fig8a}, those results are from the basal to the apical slice of two cases selected from VaC and are segmented by $Al{t_u}$, $Ne{u_u}$, $Al{t_f}$, and $Ne{u_f}$. These visual exhibitions can directly exhibit the predicted results of all networks, and we can observe the changes in segmentation results after networks incorporate CFFs (depicted as red arrows and circles). The predictions of $Al{t_u}$ could already approach the GT label maps except for some outlier predictions, whereas $Ne{u_u}$ improved those fault predictions. The results of $Al{t_f}$ and $Ne{u_f}$, respectively, which are mostly inferior to $Al{t_u}$ and $Ne{u_u}$, and the MYO predictions are less smooth. This finding suggests that the FCN network struggles to handle details, such as the border between MYO and the background, which improves after integrating CFFs. In addition, it explains why the metrics for $Al{t_f}$ and $Ne{u_f}$ are worse. Overall, the utilization of our proposed filter mechanism has led to improved segmentation results for the FCN and U-net backbone networks. 

\begin{figure}[h]
\centering
\includegraphics[width=0.9\linewidth]{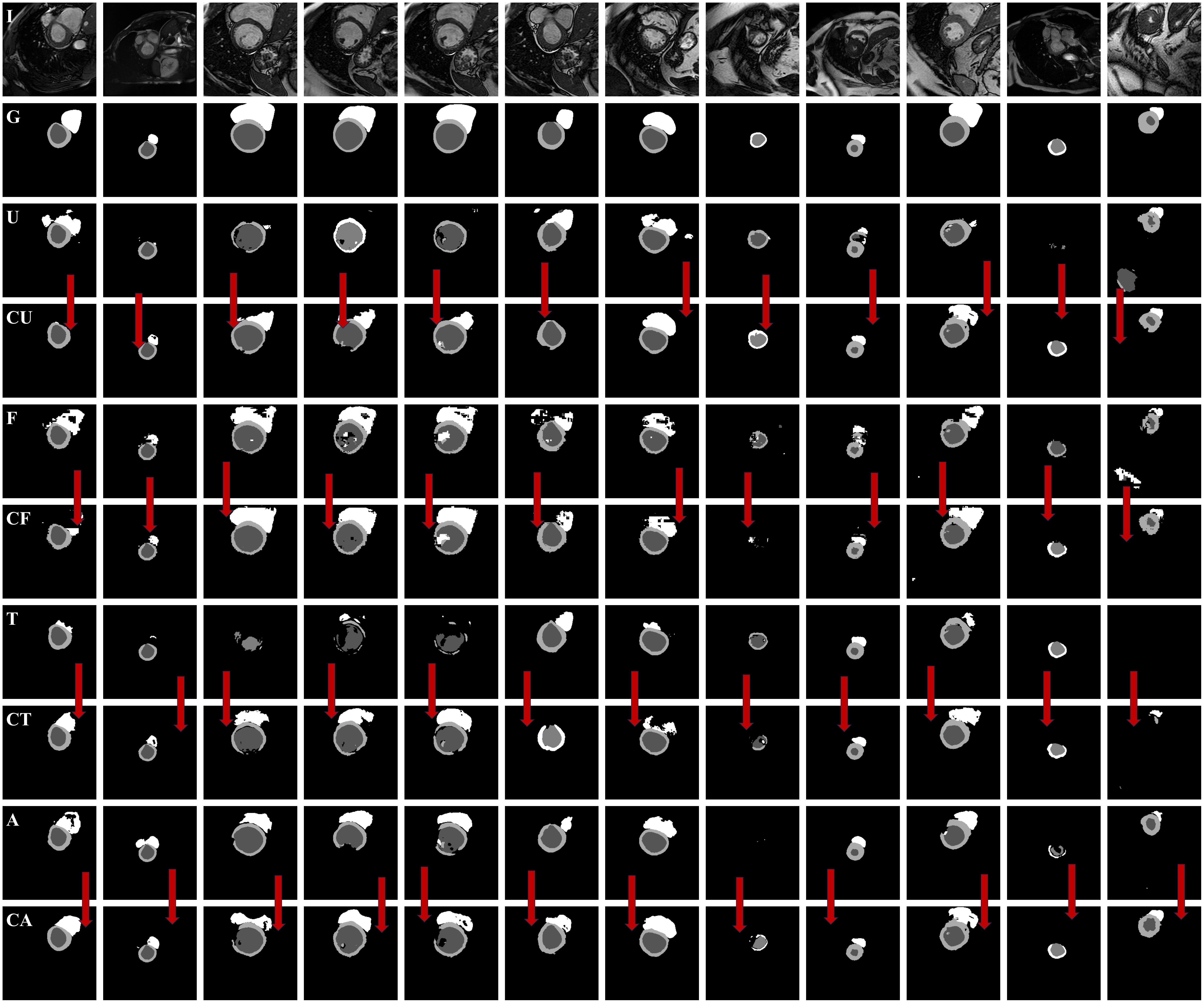}
\caption{Segmentation results on VaM obtained from different networks. The I and G are images and corresponding GT label maps, the U and CU are the results obtained from $Al{t_u}$ and $Ne{u_u}$, the F and CF are the results obtained from $Al{t_f}$ and $Ne{u_f}$, the T and CT are the results obtained from M1 and CM1, and the A and CA are the results obtained from M2 and CM2. The gray, light-gray, and white parts are LV, MYO, and RV, respectively.}
\label{fig8c}
\end{figure}

In Figure~\ref{fig8c}, the segmentation results of the eight networks on VeM are displayed. We specifically selected results from domain C and domain D, as they showcase the changes after the effective use of CFFs. While the segmentation results show improvement, they have not yet reached the level of the GT label maps, indicating the persisting challenge of domain shift, which demands further effort on our part. In addition, we observed that evaluation metrics on multi-domain data differ from those on single-domain data, with LV segmentation performing worse than RV segmentation in our experiments. This leads us to argue that the network configurations suitable for single-domain data are not well-suited for multi-domain data.

\begin{figure}[h]
\centering
\includegraphics[width=0.9\linewidth]{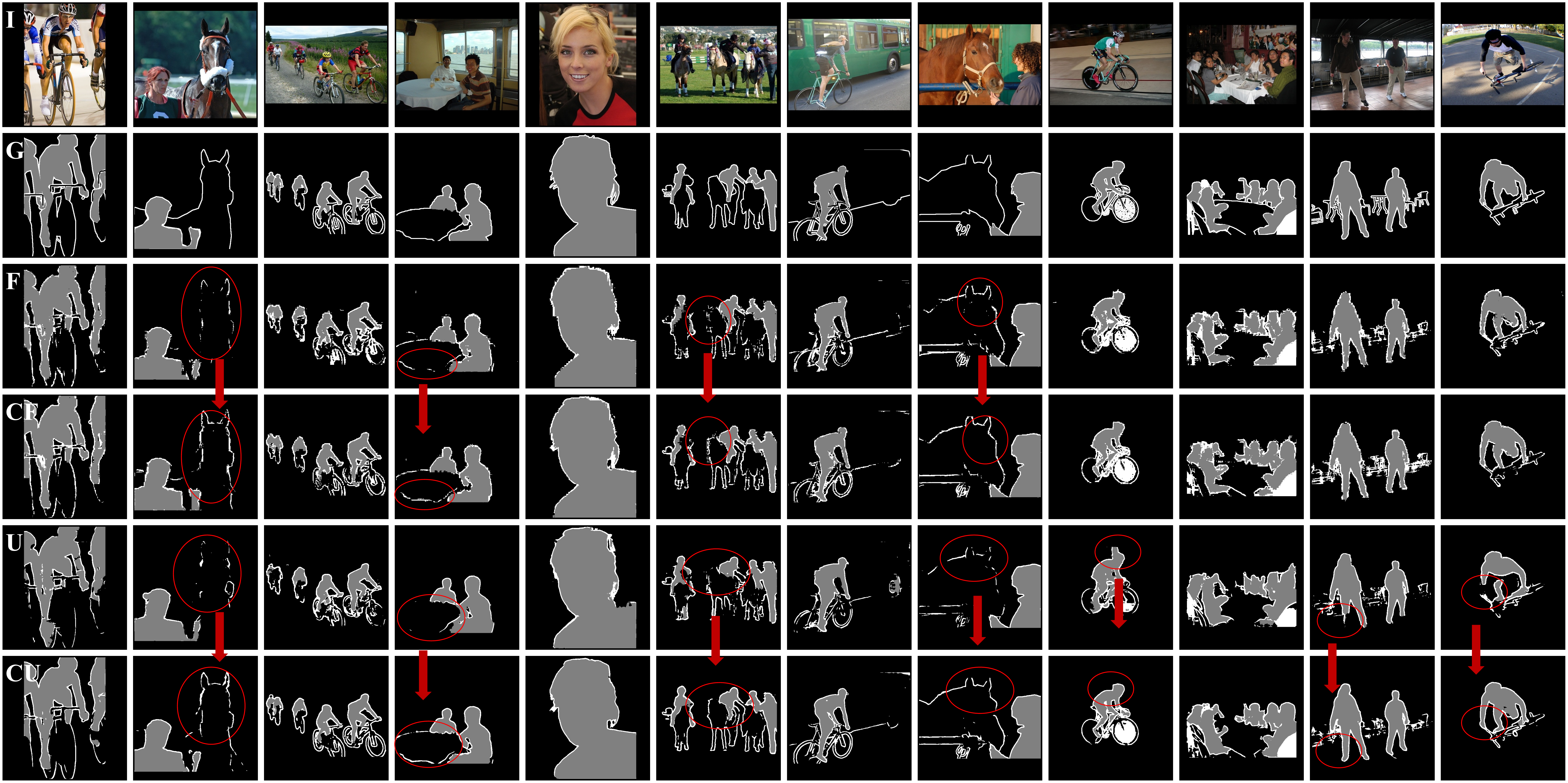}
\caption{Segmentation results on VaV obtained from different networks. The I and G are images and corresponding GT label maps, the U and CU are the results obtained from $Al{t_u}$ and $Ne{u_u}$, and the F and CF are the results obtained from $Al{t_f}$ and $Ne{u_f}$. The blue, green, and red parts are LV, MYO, and RV, respectively. The gray and white parts are humans and borders.}
\label{fig8b}
\end{figure}

Figure~\ref{fig8b} displays the results of four networks, revealing a significant improvement in BO segmentation (depicted in red arrows and circles). It is apparent that the HO segmentation results of the four networks are almost identical, suggesting that the improvement in HO segmentation is not substantial. Notably, BO predictions in the F and U rows are inferior to those in the CF and CU rows, explaining the significant improvement in BO segmentation. Intriguingly, the predicted borders of $Al{t_f}$ and $Ne{u_f}$ are still jagged, similar to the single-domain cardiac segmentation. We can conclude that the U-net excels in capturing fine details, such as thin borders, which eventually contributes to the jump concatenations in the decoder of the U-net.

Table~\ref{tab3x} shows the LV, RV, and MYO segmentation accuracy on VaC. In all sub-tasks of Table~\ref{tab3x}, $Ne{u_u}$ consistently achieved the best metrics, $Al{t_u}$ and $Ne{u_u}$ outperformed $Al{t_f}$ and $Ne{u_f}$, and $Ne{u_f}$ or $Ne{u_u}$ achieved higher scores than $Al{t_f}$ or $Al{t_u}$. Compared to the metrics of $Al{t_f}$ and $Al{t_u}$, CP1 and CP2 exhibited a few improvements. However, compared to the metrics of $Ne{u_f}$ and $Ne{u_u}$, CP1 and CP2 have some worse scores. In addition, we noticed that HD metrics have better improvements than Dice scores, and RV HD metrics are much higher than HD metrics of LV and MYO for all results. Therefore, we conclude that the segmentation ability of $Ne{u_f}$ and $Ne{u_u}$ is highly dependent on the performance of their original architecture networks. Furthermore, $Ne{u_f}$ and $Ne{u_u}$ seem to improve the segmentation results of $Al{t_f}$ and $Al{t_u}$ instead of changing them intrinsically. It attributes to the capability of CFFs, we considered, which filters noise in input feature signals.

\begin{table*}[h]
\centering
\caption{Evaluation metrics on VaC (Mean $\pm$ Std). Bold highlights indicate improvements, while red indicates the best results.}
\begin{tabular}{lcccccc}
\hline
{}&LV&{}&RV&{}&MYO&{}\\
{}&Dice (\%)&HD (mm)&Dice (\%)&HD (mm)&Dice (\%)&HD (mm)\\
\hline
$Al{t_f}$ & 90.4 $\pm$ 0.4 & 6.1 $\pm$ 0.3 & 78.7 $\pm$ 1.1 & 16.7 $\pm$ 1.3 & 84.3 $\pm$ 0.5 & 7.7 $\pm$ 0.6\\
CP1 & \textbf{90.9 $\pm$ 0.4} & \textbf{5.6 $\pm$ 0.6} & \textbf{79.8 $\pm$ 1.3} & \textbf{14.9 $\pm$ 1.1} & \textbf{84.6 $\pm$ 0.6} & \textbf{6.8 $\pm$ 0.7} \\ 
$Ne{u_f}$ & \textbf{91.0 $\pm$ 0.6} & \textbf{5.8 $\pm$ 0.5} & \textbf{80.4 $\pm$ 0.7} & \textbf{14.9 $\pm$ 1.0} & \textbf{84.6 $\pm$ 0.5} & \textbf{7.1 $\pm$ 0.4}\\ 
\hline
$Al{t_u}$ & 91.8 $\pm$ 0.6 & 4.5 $\pm$ 0.4 & 82.1 $\pm$ 0.6 & 13.1 $\pm$ 0.8 & 86.9 $\pm$ 0.5 & 5.6 $\pm$ 0.5\\ 
CP2 & \textbf{91.9 $\pm$ 0.4} & \textbf{4.3 $\pm$ 0.4} & 82.1 $\pm$ 1.5 & \textbf{12.8 $\pm$ 1.3} & \textbf{87.0 $\pm$ 0.5} &  \textcolor{red}{\textbf{4.7 $\pm$ 0.5}} \\ 
$Ne{u_u}$ &  \textcolor{red}{\textbf{92.4 $\pm$ 0.3}} &  \textcolor{red}{\textbf{3.9 $\pm$ 0.3}} &  \textcolor{red}{\textbf{82.2 $\pm$ 0.6}} & \textcolor{red}{\textbf{12.6 $\pm$ 0.8}} & \textcolor{red}{\textbf{87.6 $\pm$ 0.4}} & \textbf{4.9 $\pm$ 0.4}\\
\hline
\end{tabular}
\label{tab3x}
\end{table*}

We performed a comprehensive comparative analysis of the evaluations achieved through $Ne{u_u}$ and the state-of-the-art (SOTA) methods (see Table~\ref{tab4x}), including the top 5 methods on the ACDC leaderboard\footnote{https://www.creatis.insa-lyon.fr/Challenge/acdc/results.html} (last and final update: November 2022) and the Cross U-net (CrU) presented in August 2022. Remarkably, $Ne{u_u}$ demonstrated superior performance in terms of HD metrics, except for RV segmentation. This outcome highlights the challenging nature of RV segmentation in the context of single-domain cardiac segmentation. Notably, we merely employed the U-net backbone to achieve high-level segmentation capability. This observation emphasizes the significance of reducing feature signal noise. However, Dice scores are not significantly improved for all methods. This reason, we considered, is highly related to the ACDC data.

\begin{table}[h]
\centering
\caption{A comparison between the ACDC leaderboard, SOTA methods, and Ours (Mean $\pm$ Std, Dice (\%), HD (mm)). The red indicates the best results.}
\begin{tabular}{lcccc}
\hline
LV&Dice ED&Dice ES&HD ED&HD ES\\
\hline
G. S&96.7&\textcolor{red}{92.8}&6.4&7.6\\
F. I&96.7&92.8&5.5&6.9\\
C. Z&96.4&91.2&6.2&8.4\\
N. P&96.1&91.1&6.1&8.3\\
M. K&96.4&91.7&8.1&9.0\\
CrU&\textcolor{red}{97.0}&92.0&8.3&9.4\\
$Ne{u_u}$&96.6&92.3&\textcolor{red}{3.6}&\textcolor{red}{4.1}\\
\hline
\hline
RV&Dice ED&Dice ES&HD ED&HD ES\\
\hline
F. I&\textcolor{red}{94.6}&\textcolor{red}{90.4}&\textcolor{red}{8.8}&11.4\\
C. Z&93.4&88.5&11.0&12.6\\
N. P&93.3&88.4&13.7&13.3\\
C. Z&94.1&88.2&10.3&14.0\\
M. K&93.5&87.9&14.0&13.9\\
CrU&93.0&88.0&12.6&13.3\\ 
$Ne{u_u}$&93.8&88.2&11.6&\textcolor{red}{11.2}\\
\hline
\hline
MYO&Dice ED&Dice ES&HD ED&HD ES\\
\hline
F. I&89.6&\textcolor{red}{91.9}&7.6&7.1\\
C. Z&88.6&90.2&9.6&9.3\\
N. P&88.1&89.7&8.6&9.6\\
M. K&88.9&89.8&9.8&12.6\\
S. J&88.2&89.7&9.8&11.3\\
CrU&89.0&90.0&10.0&10.7\\
$Ne{u_u}$&\textcolor{red}{89.7}&91.2&\textcolor{red}{5.1}&\textcolor{red}{5.6}\\
\hline
\end{tabular}
\label{tab4x}
\end{table}

Table~\ref{tab5x} presents the evaluation results of all networks on VaM. First, when analyzing the metrics of the networks without CFFs, we observed that the performance of U-net-based networks surpassed that of FCN, consistent with the results on single-domain data. Furthermore, the transformer and CABM techniques notably improved the segmentation ability of the U-net backbone. Second, upon analyzing the metrics of the networks with CFFs, we found significant improvements in both Dice and HD metrics compared to the previous results. Specifically, CM2 achieved the best Dice scores, while $Ne{u_u}$ performed the best in terms of HD metrics. These findings indicate that our proposed CFFs are effective even when dealing with multi-domain data. 

\begin{table}[h]
\centering
\caption{Evaluation metrics on VaM (Mean $\pm$ Std). Bold highlights indicate improvements, while red indicates the best results.}
\begin{tabular}{lcc}
\hline
LV&Dice (\%)&HD (mm)\\
\hline
$Al{t_f}$ & 79.6 $\pm$ 3.7 & 23.6 $\pm$ 7.8 \\
$Ne{u_f}$ & \textbf{83.0 $\pm$ 0.7} & \textbf{18.4 $\pm$ 1.3} \\
\hline
$Al{t_u}$ & 82.7 $\pm$ 1.7 & 18.0 $\pm$ 3.0 \\ 
$Ne{u_u}$ & \textbf{84.5 $\pm$ 1.6} & \textbf{15.1 $\pm$ 1.7} \\
\hline
M1 & 82.8 $\pm$ 0.8 & 17.4 $\pm$ 1.5 \\
CM1 & \textbf{84.2 $\pm$ 1.5} & \textbf{15.3 $\pm$ 1.8} \\
\hline
M2 & 81.7 $\pm$ 3.0 & 18.3 $\pm$ 3.9 \\
CM2 & \textcolor{red}{\textbf{85.3 $\pm$ 1.7}} & \textcolor{red}{\textbf{14.7 $\pm$ 2.0}} \\ 
\hline
\hline
RV&Dice (\%)&HD (mm)\\
\hline
$Al{t_f}$ & 83.6 $\pm$ 2.3 & 20.5 $\pm$ 5.0 \\
$Ne{u_f}$ & \textbf{86.3 $\pm$ 0.7} & \textbf{15.9 $\pm$ 0.8} \\
\hline
$Al{t_u}$ & 85.7 $\pm$ 1.0 & 16.6 $\pm$ 1.6 \\ 
$Ne{u_u}$ & \textbf{87.8 $\pm$ 1.2} & \textbf{13.1 $\pm$ 2.2} \\
\hline
M1 & 85.9 $\pm$ 0.7 & 16.3 $\pm$ 1.8 \\
CM1 & \textbf{86.8 $\pm$ 1.1} & \textbf{15.2 $\pm$ 1.8} \\
\hline
M2 & 85.8 $\pm$ 2.5 & 16.4 $\pm$ 4.1 \\
CM2 & \textcolor{red}{\textbf{87.9 $\pm$ 0.1}} & \textcolor{red}{\textbf{13.1 $\pm$ 0.7}} \\
\hline
\hline
MYO&Dice (\%)&HD (mm)\\
\hline
$Al{t_f}$ & 77.2 $\pm$ 2.5 & 22.7 $\pm$ 4.8 \\
$Ne{u_f}$ & \textbf{79.8 $\pm$ 1.0} & \textbf{19.0 $\pm$ 1.6} \\
\hline
$Al{t_u}$ & 79.2 $\pm$ 1.5 & 18.4 $\pm$ 2.5 \\ 
$Ne{u_u}$ & \textbf{82.4 $\pm$ 1.0} & \textcolor{red}{\textbf{14.9 $\pm$ 1.8}} \\
\hline
M1 & 80.0 $\pm$ 1.3 & 18.7 $\pm$ 2.6 \\
CM1 & \textbf{81.4 $\pm$ 1.1} & \textbf{17.2 $\pm$ 1.9} \\
\hline
M2 & 79.1 $\pm$ 2.9 & 18.9 $\pm$ 3.8 \\
CM2 & \textcolor{red}{\textbf{82.5 $\pm$ 0.5}} & \textbf{15.0 $\pm$ 1.0} \\
\hline
\end{tabular}
\label{tab5x}
\end{table}

\subsection{Others}

In this presentation, we provide several specific details to enhance the comprehension of our research efforts.  

1. A low anti-noise property of convolution.

\textit{Proof} Let's consider a signal with noise, denoted by $\alpha {x_n} + \beta {\varepsilon _n}$, where $\alpha$ and $\beta$ are constants and satisfy the condition $\alpha + \beta = 1$. The convolution operation of this signal with a function $G(k)$ can be linearly decomposed as follows: $G(k) * (\alpha {x_n} + \beta {\varepsilon _n}) = \alpha G(k) * {x_n} + \beta G(k) * {\varepsilon _n}$. This linear decomposition reveals that the noise persists even after convolution, indicating that the noise consistently affects the subsequent operations.

2. Differences between CFFs and AMs.

While the mechanism of a CFF may appear similar to Attention Mechanisms (AMs), they differ in terms of conception and the service object. AMs are associated with self-adaptive weights for convolutional features, whereas a CFF is a natural-selected feature optimization method. AMs typically refer to one vector that corresponds to multiple channel-, spatial-, or temporal-based features (one-to-many). In contrast, a CFF involves one filter matrix, the output from $s(conv(f,\theta ))$, that corresponds to a single input feature signal matrix (one-to-one). 

3. Data Motivation and Challenges. 

The motivation for conducting experiments on these three datasets is as follows: For the short-axis cardiac MR image segmentation task, the ACDC and M\&Ms datasets serve as standard research data, readily accessible at no cost. The VOCH dataset, on the other hand, is intended solely to validate the performance of CFFs in non-medical segmentation, thus serving as a reference task.

The ACDC dataset presented challenging characteristics, including varying shapes of segmentation targets and difficulties in distinguishing them from the backgrounds. For instance, the right ventricle exhibited significant shape changes across different slices, and the myocardium, liver, and left lung shared the same signal intensity. Both the training and testing data belonged to a single domain with the same distribution. The M\&Ms dataset shared the challenges of the ACDC data, with an additional significant issue of domain shift problems. Its primary objective was to advance research and establish scientific benchmarks for generalizable deep learning in cardiac segmentation. The training data comprised three known domains, while the testing data included one unseen domain. For achieving better segmentation results, a strong emphasis was placed on the generalization ability of networks. The VOCH dataset similarly featured challenging characteristics, such as diverse backgrounds and varying postures of segmentation targets. For instance, subjects could be pictured riding a bike outdoors or sleeping in a bedroom. The training and validation data of VOCH belonged to a single domain with the same distribution.

\end{document}